\documentclass[10pt,twocolumn,letterpaper]{article}

\usepackage{cvpr}
\usepackage{times}
\usepackage{epsfig}
\usepackage{graphicx}
\usepackage{amsmath}
\usepackage{amssymb}
\usepackage{caption,subfig}
\usepackage{float}
\usepackage{xcolor}
\usepackage{tikz-qtree}
\usepackage{tikz}
\usepackage[draft]{todonotes}   % notes showed
\usepackage[pagebackref=true,breaklinks=true,letterpaper=true,colorlinks,bookmarks=false]{hyperref}

\usepackage{algorithm}
\usepackage{multirow}
\usepackage{mathtools,xparse}
\usepackage{siunitx}
\usepackage{stmaryrd}
\usepackage[noend]{algpseudocode}
\usepackage{placeins}
\usepackage[export]{adjustbox}
\usepackage{diagbox}

\usetikzlibrary{trees} % this is to allow the fork right path
\usetikzlibrary{shapes,arrows,decorations.markings}
\usetikzlibrary{spy}
\usetikzlibrary{calc}
\usetikzlibrary{intersections}

\cvprfinalcopy % *** Uncomment this line for the final submission

\def\cvprPaperID{2157} % *** Enter the CVPR Paper ID here

\newcommand{\col}{\mu}
\newcommand{\rad}{r}

\newcommand{\traj}[1]{P_{#1}}
\newcommand{\frm}[1]{I_{#1}}

\newcommand{\degop}{\mathcal{H}}
\newcommand{\expFrac}{\varepsilon}
\newcommand{\datasetScale}{0.245}
\newcommand{\imgZoom}[4]{
\begin{tikzpicture}[spy using outlines={circle,yellow,magnification=#3,size=1.5cm, connect spies}]
    \node[outer sep=0pt, inner sep=0pt] (image) { \includegraphics[width=\datasetScale\textwidth]{#1} };
 	\begin{scope}[x={(image.south east)},y={(image.north west)}]
    	\spy on #2 in node [left] at #4;
    \end{scope}
\end{tikzpicture}
}

\newcommand{\imgZoomIntro}[4]{
\begin{tikzpicture}[spy using outlines={circle,yellow,magnification=#3,size=1.5cm, connect spies}]
    \node[outer sep=0pt, inner sep=0pt] (image) { \includegraphics[width=0.49\linewidth]{#1} };
    \begin{scope}[x={(image.south east)},y={(image.north west)}]
        \spy on #2 in node [left] at #4;
    \end{scope}
\end{tikzpicture}
}

\tikzset{cross/.style={cross out, draw, minimum size=2*(#1-\pgflinewidth), inner sep=0pt, outer sep=0pt}}
\newcommand{\drawImageAndCenters}[8]
{
\begin{tikzpicture}
    \node[anchor=south west,inner sep=0] (image) at (0,0) {\includegraphics[width=#2]{#1}};
    \begin{scope}[x={(image.south east)},y={(image.north west)}]
        \draw[red, ultra thick] (#5/#3,{(#4-#6)/#4}) circle (0.10cm);
        \draw (#7/#3,{(#4-#8)/#4}) node[cross=0.14cm,rotate=0,blue, ultra thick]{};
    \end{scope}
\end{tikzpicture}
}
\newcommand{\drawImage}[2]
{
\begin{tikzpicture}
    \node[anchor=south west,inner sep=0] (image) at (0,0) {\includegraphics[width=#2]{#1}};
\end{tikzpicture}
}

\newcommand{\cscl}{}
\newcommand{\defaultRight}{(2,-0.4)}
\newcommand{\defaultLeft}{(-0.5,-0.4)}
\newcommand{\ImgSpace}{15pt}
\newcommand{\LblAbove}{-6pt}
\newcommand{\LblRight}{-27pt}

\newcommand{\volleyballCoor}{(0.83, -0.57)}
\newcommand{\volleyballpassingCoor}{(0.42, 0.26)}
\newcommand{\dartsCoor}{(-0.69, 0.41)}
\newcommand{\dartswindowCoor}{(1.32, 0.26)}
\newcommand{\softballCoor}{(-1.03, 0.48)}
\newcommand{\williamtellCoor}{(0.37, 0.59)}
\newcommand{\tennisservesideCoor}{(1.02, 0.94)}
\newcommand{\tennisservebackCoor}{(-0.24, 0.49)}
\newcommand{\tennisCoor}{(-0.29, 0.22)}
\newcommand{\hockeyCoor}{(0.82, 0.11)}
\newcommand{\squashCoor}{(-0.47, 0.26)}
\newcommand{\frisbeeCoor}{(-0.81, 0.0292)}
\newcommand{\blueCoor}{(-0.0836, 0.37)}
\newcommand{\pingpongpaintCoor}{(-1.38, 0.0477)}
\newcommand{\pingpongsideCoor}{(0.57, 0.03)}
\newcommand{\pingpongtopCoor}{(0.04, 0.55)}

\DeclareGraphicsExtensions{
	.jpg,
    .png,.PNG,%
    .pdf,.PDF,%
    .mps,.jpeg,.jbig2,.jb2,.JPG,.JPEG,.JBIG2,.JB2}
\graphicspath{{./img/}}

\ifcvprfinal\fi
\begin{document}

%%%%%%%%% TITLE
\title{The World of Fast Moving Objects}

\author{Denys Rozumnyi$^{1,3}$ \\
% {\tt\small rozumden@cmp.felk.cvut.cz}
\and
Jan Kot\v{e}ra$^{2}$\\
% {\tt\small kotera@utia.cas.cz}
\and
Filip \v{S}roubek$^{2}$ \\
% {\tt\small sroubekf@utia.cas.cz}
\and
Luk\'a\v{s} Novotn\'y$^{1}$\\
% {\tt\small novotl22@fel.cvut.cz}
\and
Ji\v{r}\'i Matas$^{1}$\\
% {\tt\small matas@cmp.felk.cvut.cz}
\and
% $^{1}$Center for Machine Perception, Czech Technical University in Prague 
$^{1}$CMP, Czech Technical University in Prague 
\and
% $^{2}$Institute of Information Theory and Automation, Czech Academy of Sciences
$^{2}$UTIA, Czech Academy of Sciences
\and
% $^{3}$Department of Signal Processing, Tampere University of Technology
$^{3}$SGN, Tampere University of Technology
}

\maketitle
%\thispagestyle{empty}

%%%%%%%%% ABSTRACT
\begin{abstract}
   The notion of a Fast Moving Object (FMO), i.e. an object that moves over a distance exceeding its size within the exposure time, is introduced. FMOs  may, and typically do, rotate with high angular speed. FMOs are very common in sports videos, but are not rare elsewhere. In a single frame, such objects are often barely visible and appear as semi-transparent streaks.

A method for the detection and tracking of FMOs is proposed. The method consists of three distinct algorithms, which form an efficient localization pipeline that operates successfully in a broad range of conditions. We show that it is possible to recover the appearance of the object and its axis of rotation, despite its blurred appearance. The proposed method is evaluated on a new annotated dataset. The results show that existing trackers are inadequate for the problem of FMO localization and a new approach is required. Two applications of localization, temporal super-resolution and highlighting, are presented.
%Three applications of such localization, temporal super-resolution, highlighting and background inpainting, are presented.   
\end{abstract}

%%%%%%%%% BODY TEXT
\section{Introduction} \label{sec:intro}
\begin{figure}
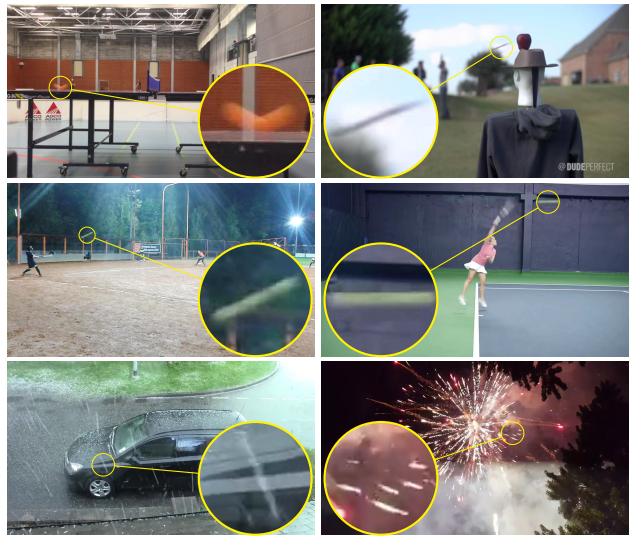

\centering
\begin{tabular}{@{}c@{}c@{}}
	\imgZoomIntro{seq/ping_pong_paint}{(-1.34, 0.0477)}{5}{\defaultRight} &
	\imgZoomIntro{seq/william_tell}{(0.35, 0.58)}{5}{\defaultLeft} \\

	\imgZoomIntro{seq/softball}{(-0.985, 0.46)}{7}{\defaultRight} &
	\imgZoomIntro{seq/tennis_serve_side}{(0.97, 0.90)}{5}{\defaultLeft} \\

	\imgZoomIntro{seq/hail}{(-0.77, -0.22)}{5}{\defaultRight} &
	\imgZoomIntro{seq/fireworks}{(0.5,0.2)}{5}{\defaultLeft} \\
\end{tabular}
\caption{Fast moving objects appear as semi-transparent streaks larger than their size. Examples (left-to-right, top-to-bottom) from table tennis, archery, softball, tennis, hailstorm and fireworks.}
\label{fig:intro}
\end{figure}

Object tracking has received enormous attention by the computer vision community. Methods based on various principles have been proposed and several surveys have been compiled \cite{Avidan:2007:ET:1191552.1191804,5674053,Godec:2013:HTN:2527401.2527616}. Standard benchmarks, some comprising hundreds of videos, such as ALOV \cite{alov}, VOT \cite{vot2015,vot2016} and OTB \cite{otb} are available. Yet none of them include objects that are moving so fast that they appear as streaks much larger than their size. This is a surprising omission considering the fact that such objects commonly appear in diverse real-world situations, in which sports play undoubtedly a prominent role; see examples in Fig. \ref{fig:intro}%
\footnote{Fast moving objects are often poorly visible and for improved understanding, the reader is referred to videos in the supplementary material.}.

\begin{figure} [t]
\begin{tabular}{cc}
 \includegraphics[width=0.46\linewidth]{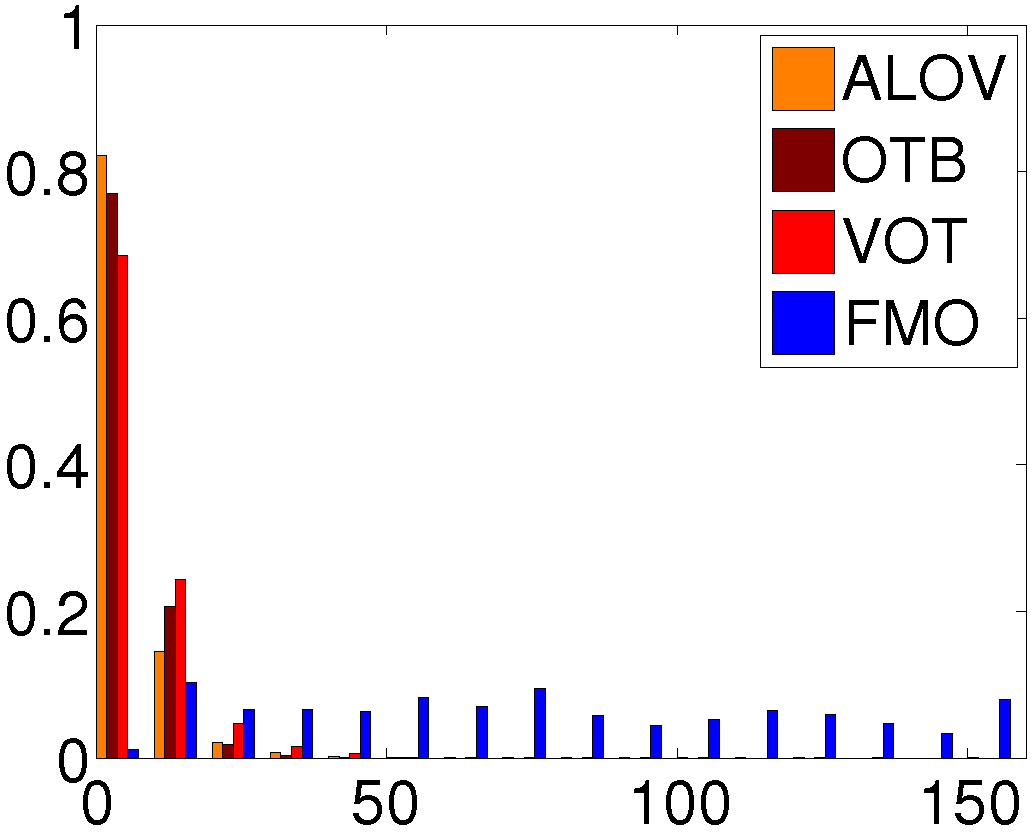} & 
 \includegraphics[width=0.46\linewidth]{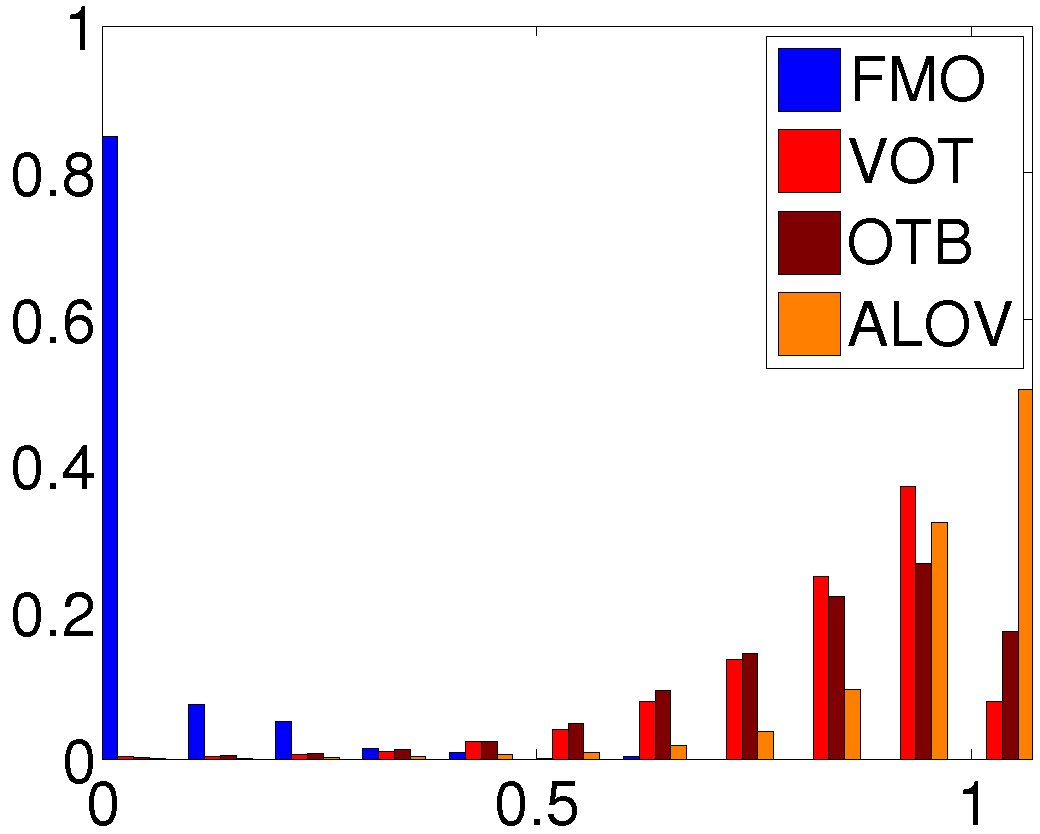} \\
 object speed $[\text{pxl}]$ & intersection over union 
\end{tabular}
\caption{The FMO dataset includes motions that are an order of magnitude faster than three standard datasets - ALOV, VOT, OTB \cite{alov,vot2015,otb}. Normalized histograms of projected object speeds (left) and intersection over union IoU of bounding boxes (right) between adjacent frames.}
\label{fig:hist}
\end{figure}

\def\arraystretch{0.5}
\def\arraystretch{1} 
To develop 	algorithms for detection and tracking of fast moving objects, we had to collect and annotate a new dataset.  The substantial difference of the FMO dataset and the standard ones was confirmed by ex-post analysis of inter-frame motion statistics. The most common overlap of ground truth bounding boxes in two consecutive frames is {\it zero} for the FMO set while it is {\it close to one} for ALOV, OTB and VOT \cite{alov,otb,vot2015}. The speed of the tracked object projected to image coordinates, measured as the distance of object centers in two consecutive frames, is on average ten times higher in the new dataset,  see Fig. \ref{fig:hist}. Given the difference in the properties of the sequences, it is not surprising that state-of-the-art trackers designed for the classical problem do not perform well on the FMO dataset. The two ``worlds'' are so different that  on almost all sequences the classical state-of-the-art methods fail completely, their output bounding boxes achieving a 50\% overlap with the ground truth in  {\it zero} frames, see Tab. \ref{tab:baselines}.

In the paper, we propose an efficient method for FMO localization and a method for estimation of their extrinsic -- the trajectory and the axis and angular velocity of rotation, and intrinsic properties --  the size and color of the object. In specific cases we can go further and estimate the full appearance model of the object. Properties like the rotation axis, angular velocity and object appearance require precise modeling of the image formation (acquisition) process.  The proposed method thus proceeds by solving a blind space-variant deconvolution problem with occlusion.  

%We further demonstrate that FMOs enable estimating some properties of the image sensor, \eg the exposure fraction or the type of the sensor -- rolling shutter, progressive or interlaced scan. 
%%Denis - we don't have any experiments with the type of sensor%%
%
Detection, tracking and appearance reconstruction of FMOs allows performing tasks with applications in diverse areas. We show, for instance, the ability to synthesize realistic videos with higher frame rates, i.e. to perform temporal super-resolution. The extracted properties of the FMO, such as trajectory, rotation angle and velocity have application, \eg in sports analytics. 

The rest of the paper is organized as follows: Related work is discussed in Section \ref{sec:related}. Section \ref{sec:problem} defines the main concepts arising in the problem.  Section \ref{sec:method} explains in detail the proposed method for FMO localization.  The estimation of intrinsic and extrinsic properties formulated as an optimization problem is presented in Section \ref{sec:deconv}. In Section~\ref{sec:dataset}, the FMO annotated dataset of $16$ videos is introduced. Last, the method is evaluated and its different applications are demonstrated in Section \ref{sec:eval}.

\section{Related Work}  \label{sec:related}
Tracking is a key problem in video processing. A range of methods has been proposed based on diverse principles, such as correlation \cite{Biresaw:2015:CST:2729305.2741247, dsst,srdcf}, feature point tracking \cite{tomasi1991detection}, mean-shift \cite{Comaniciu:2003:KOT:776753.776799,asms},  and tracking-by-detection \cite{meem,struck}. The literature sometimes refer to fast moving objects, but only the case with no significant blur is considered, \eg \cite{zaveri,Kruglov2016}.

Object blur is a cue for object motion, since the blur size and shape encode information about motion. However, classical tracking methods suffer from blur, yet FMOs consist predominantly of blur. 
%, i.e., in every neighborhood it is defined by the angle and the length of line blur. 
%If some additional prior information about the object motion is available, for example that the object rotates, it helps to constrain further the local linear model and alleviates the ill-posed nature of such space-variant blind deconvolution problems.     
%
Most motion deblurring methods assume that the degradation can be modeled locally by a linear motion. One category of methods works with occlusion and considers the object's blurred transparency map \cite{Jia2007}. Blind deconvolution of the transparency map is easier, since the latent sharp map is a binary image. The same idea applied to rotating objects was proposed in \cite{Shan2007}. An interesting variation was proposed in \cite{daiwu:08}, where linear motion blur is estimated locally using a relation similar to optical flow. The main drawback of these methods is that an accurate estimation of the transparency map using alpha matting algorithms \cite{Levin2008} is necessary.
%work only for objects with relatively small motion blur. 
%\cite{Almeida2009}

Methods exploiting the fact that autocorrelation increases in the direction of blur were proposed to deal with objects moving over static backgrounds \cite{Chakrabarti2010,Kim2014}. 
%The blur length is proportional to the autocorrelation extent, and unlike the blur direction, it is more difficult to estimate accurately.  
Similarly \cite{Oliveira2014, Sun2015}, autocorrelation was considered for  motion detection of the whole scene due to camera motion. However, all these methods require a relatively large neighborhood to estimate blur parameters, which means that they are not suitable for small moving objects. Simultaneously dealing with rotation of objects has not been considered in the literature so far. 

\section{Problem definition}  \label{sec:problem}
FMOs are objects that move over a large distance compared to their size during the exposure time of a single frame, and possibly also rotate along an arbitrary axis with an unknown angular speed. For simplicity, we assume a single object $F$ moving over a static background $B$; an extension to multiple objects is relatively straightforward. 
To get close to the static background state, camera motion is assumed to be compensated by video stabilization. 

Let a recorded video sequence consist of frames $\frm{1}(x),\ldots \frm{n}(x)$, where $x \in \mathbb{R}^2$ is a pixel coordinate. 
%For brevity, we omit the position argument $x$ when it is not needed. 
Frame $\frm{t}$ is formed  as
\begin{equation}
%\frm{t} = (1-\degop_t[M])B + \degop_t[F]\,,
\frm{t}(x) = (1-[\degop_t M](x)) B(x) + [\degop_t F](x)\,,
\label{eq:model}
\end{equation}
where $M$ is the indicator function of $F$. In general, the operator $\degop_t$ models the blur caused by object motion and rotation, and performs the 3D$\to$2D projection of the object representation $F$ onto the image plane. This operator depends mainly on three parameters, $\{\traj{t},a_t,\phi_t\}$, which are the FMO trajectory (path), and the axis and angle of rotation, respectively.
%Note that in the general case, $F$ represents an unknown 3D object and then $\degop_t[F]$ also performs projection into the image plane.
The  $[\degop_t M](x)$ function corresponds to the object visibility map (alpha matte, relative duration of object presence during exposure) and appears in \eqref{eq:model} to merge the blurred object and the partially visible background.

The object trajectory $\traj{t}$ can be represented in the image plane as a path (set of pixels) along which the object moves during the frame exposure. In the case of no rotation or when $F$ is homogeneous, i.e. the surface is uniform and thus rotation is not perceivable, $\degop_t$ simplifies to a convolution in the image plane, i.e. $[\degop_t F](x) = \frac{1}{|\traj{t}|}[\traj{t}\ast F](x)$, where $|\traj{t}|$ is the path length~--~$F$ can then be viewed as a 2D image.

Finding all the intrinsic and extrinsic properties of arbitrary FMOs means estimating both $F$ and $\degop_t$, which is, at this moment, an intractable task. To alleviate this problem, some prior knowledge of $F$ is necessary. In our case, the prior is in the form of object shape. Since in most sport videos the FMOs are spheres (balls), we continue our theoretical analysis focusing on spherical objects, although, as we further demonstrate, the proposed localization method can also successfully handle objects of significantly different shapes.

We propose methods for two tasks: (i) efficient and reliable FMO localization, i.e. detection and tracking, and (ii) reconstruction of the FMO appearance, and the axis and angle of the object rotation, which requires the precise output of (i).  For tracking, we use a simplified version of \eqref{eq:model} and approximate the FMO by a homogeneous circle determined by two parameters: color $\col$ and radius $\rad$. The tracker output (trajectory $\traj{t}$ and radius $\rad$) is then used to initialize the precise estimation of appearance using the full model \eqref{eq:model}.

\newcommand{\examplesHeight}{0.165}
\newcommand{\examplesSpacing}{\hskip 4px}
\begin{figure*}
\centering
\begin{tabular}{@{}c@{}c@{\examplesSpacing}c@{}c@{\examplesSpacing}c@{}c@{}}
    \includegraphics[frame,height=\examplesHeight\textwidth]{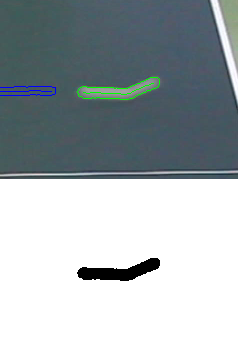} & 
    \includegraphics[frame,height=\examplesHeight\textwidth]{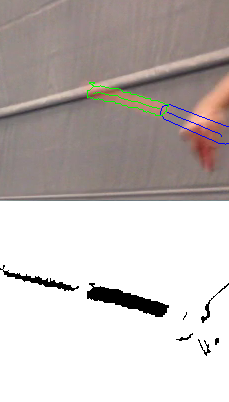} &
    \includegraphics[frame,height=\examplesHeight\textwidth]{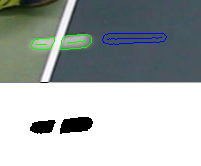} &
    \includegraphics[frame,height=\examplesHeight\textwidth]{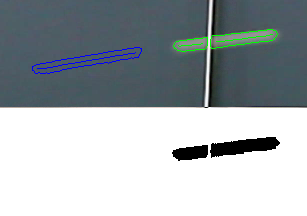} &
    \includegraphics[frame,height=\examplesHeight\textwidth]{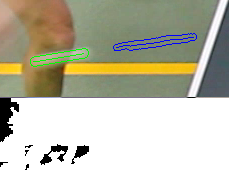} &
    \includegraphics[frame,height=\examplesHeight\textwidth]{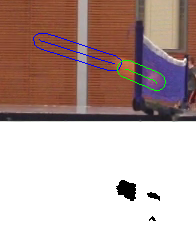} \\
    \multicolumn{2}{c}{(a) Detection} & 
    \multicolumn{2}{c}{(b) Re-detection} & 
    \multicolumn{2}{c}{(c) Tracking} \\ 
\end{tabular}
\caption{
(a) FMO detection, (b) redetection where detection failed because FMO is not a single connected component, (c) tracking where both algorithms failed due to imprecise $\Delta$. Top row: cropped $\frm{t}$ with $\traj{t-1}$ (blue) and $\traj{t}$ (green) with contours.
Bottom row: binary differential image $\Delta$. 
}
\label{fig:examples}
\end{figure*}
\section{Localization of FMOs} \label{sec:method}
The proposed FMO localization pipeline consists of three algorithms that differ in their prerequisites and speed. First, the pipeline runs the fastest algorithm and terminates if a fast moving object is localized; otherwise, 
%a more complex and general algorithm is executed. 
it proceeds to run the remaining two more complex and general algorithms. 
This strategy produces an efficient localization method that operates successfully in a broader range of conditions than either of the three algorithms alone. We call them detector, re-detector, and tracker, and their basic properties are outlined in Tab. \ref{tab:pipeline}. 
 
The first algorithm, detector, discovers previously unseen FMOs and establishes their properties. It requires sufficient contrast between the object and the background, an unoccluded view in three consecutive frames, and no interference with other moving objects. Then the FMO can be tracked by either of the other two algorithms. The second algorithm, re-detector, is applied in a region predicted by the FMO trajectory in the previous frames. It handles the problems of partial occlusions and object-background appearance similarity while being as fast as the detector. Finally, the tracker searches for the object by synthesizing its appearance with the background at the predicted locations.

\begin{table} [t]
\begin{tabular}{c||c|c|c}
 & Detector & Redetector & Tracker \\
\hline
\multirow{2}{*}{IN} & \multirow{2}{*}{$\frm{t-1}$, $\frm{t}$, $\frm{t+1}$} & $\frm{t-1}$, $\frm{t}$, $\frm{t+1}$ & $\frm{t}$, $\expFrac$ \\
 &  & $\col$, $\rad$, $\traj{t-1}$, $\expFrac$ & $\col$, $\rad$, $\traj{t-1}$ \\
\hline
\multirow{2}{*}{OUT} & \multirow{2}{*}{$\col$, $\rad$, $\traj{t}$} & \multirow{2}{*}{$\traj{t}$} & \multirow{2}{*}{$\traj{t}$} \\
 &  &  & \\
\hline
\multirow{5}{*}{ASM} & high contrast, & high contrast, & linear traj., \\
 & fast movement, & fast movement, & model \\
 & no contact with & model & \\
 & moving objects, & & \\
 & no occlusion, & & \\
\end{tabular}
\caption{Inputs, outputs and assumptions of each algorithm. Image frame at time $t$ is denoted by $\frm{t}$. Symbols $\col$ and $\rad$ are used for FMO intrinsics -- mean color and radius. FMO trajectory in $\frm{t}$ is marked by $\traj{t}$, camera exposure fraction by $\expFrac$.}
\label{tab:pipeline}
\end{table}

% \begin{figure} [H]
% \begin{tabular}{c|c|c}
% Name & Inputs & Assumptions \\
% \hline
% \hline
% \multirow{2}{*}{Detector} & \multirow{2}{*}{$\frm{t-1}$, $\frm{t}$, $\frm{t+1}$} & distinct $B$, fast movement, \\
%  & & no contact and occlusions \\
%  \hline
% \multirow{2}{*}{Redetector} & $\frm{t-1}$, $\frm{t}$, $\frm{t+1}$ & partially distinct $B$,  \\
%  &  $\col$, $\rad$, $\traj{t-1}$ & fast movement, model based \\
% \hline
% \multirow{2}{*}{Tracker} & $\frm{t}$, $\expFrac$ & linear trajectory, \\
%  &  $\col$, $\rad$, $\traj{t-1}$ & model based \\
% \end{tabular}
% \caption{Inputs and assumptions of each method.}
% \label{fig:pipeline}
% \end{figure}

All three algorithms require a static background or a registration of consecutive frames. To this end, we apply video stabilization by estimating the affine transformation between frames using RANSAC \cite{ransac} by matching FREAK descriptors \cite{freak} of FAST features \cite{fast}.

The detector also updates the FMO model properties required by the re-detector and tracker, namely FMO's color $\col$ and radius $\rad$. For increased stability, the new value of any of these parameters is a weighted average of the detected value and the previous value using a forgetting function proposed in \cite{forgetting}. For each video sequence we also need to determine the so called exposure fraction $\expFrac$, which is the ratio of exposure period and time difference between consecutive frames (e.g. 25fps video with 1/50s exposure has $\expFrac=0.5$). This can be done from any two subsequent FMO detections and we use average over multiple observations. 

%Real-time performance is a necessity in many video processing tasks. Our MATLAB implementation achieves this speed if running the detector and/or re-detector suffices for FMO localization. The third method (tracker) is more complex and therefore slower. Its principle is not very suitable for implementation in MATLAB and quite possibly a C implementation would improve the performance.
%Also, when only detector/re-detector is used, the background stabilization non-negligibly contributes to the overall runtime~--~we can achieve super real-time performance for sequences where this step is not necessary.
We need three consecutive video frames to localize the FMO in the second of the three frames, which causes a constant delay of one frame in real-time processing, but this does not present any obstacle for practical use.

\subsection{Detector}
The detector is the only generic algorithm for FMO localization that requires no input, except for three consecutive image frames $\frm{t-1}, \frm{t}, \frm{t+1}$.
%Detector is a generic method for FMO localization. It requires 3 consecutive frames $(\frm{t-1}, \frm{t}, \frm{t+1})$ to localize the FMO in $\frm{t}$ but does not need any other input, like model parameters or prediction from previous detections.
First we compute differential images $\Delta_+ = \left|\frm{t} - \frm{t-1}\right|$, $\Delta_0 = \left|\frm{t+1} - \frm{t-1}\right|$, and $\Delta_- = \left|\frm{t} - \frm{t+1}\right|$. These are binarized (denoted by superscript $b$) by thresholding, and the resulting images are combined by a boolean operation to a single binary image
\begin{equation}
\Delta = \Delta_+^b \wedge \Delta_-^b \wedge \neg \Delta_0^b.
\label{eq:deltasToFmo}
\end{equation}
This image contains all objects, which are present in the frame $\frm{t}$, but not in the frames $\frm{t-1}$ and $\frm{t+1}$ (i.e. moving objects in $\frm{t}$).

The second step is to identify all objects which can be explained by the FMO motion model. We calculate the trajectory $\traj{t}$ and radius $\rad$ for each FMO candidate and determine if it satisfies the motion model. For each connected component $C$ in $\Delta$, we compute the distance transform to get the minimal distance $d(x)$ for each inner pixel $x\in C$ to a pixel on its component's contour. Then the maximum of such distances for each component is its radius, $\rad=\max d(x),\ x\in C$. Next, we determine the trajectory by morphologically  thinning the pixels $x$ that satisfy $d(x) > \psi \rad$, where the threshold $\psi$ is set to $0.7$. Now we decide whether the object satisfies the FMO motion model by verifying two conditions: (i) the trajectory $\traj{t}$ must be a single connected stroke, and (ii) the area $a$ covered by the component $C$ must correspond to the area $\hat{a}$ expected according to the motion model, that is $\hat{a} = 2 \rad \left| \traj{t} \right| + \pi \rad ^2$. %
% \begin{equation}
% \hat{a} = 2 \rad \left| P_t \right| + \pi \rad ^2.
% \label{eq:exp_area}
% \end{equation}
We say that the areas correspond, if $\left| \frac{a}{\hat{a}} - 1 \right| < \gamma$, where $\gamma$ is a chosen threshold $0.2$. All components which satisfy these two conditions are then marked as FMOs. The whole algorithm is pictorially described in Fig.~\ref{fig:det}.

\subsection{Re-detector}
The re-detector requires the knowledge of the FMO, but allows one FMO occurrence to be composed of several connected components in $\Delta$ (e.g. the FMO passes in front of background with similar color). Fig.~\ref{fig:examples} shows an example, where the re-detector finds an FMO missed by the detector. 

The re-detector operates on a rectangular window of a binary differential image $\Delta$ in \eqref{eq:deltasToFmo}, restricted to the local neighborhood of the previous FMO localization. Let $\traj{t-1}$ be the trajectory from the previous frame $\frm{t-1}$, then the re-detector works in the square neighborhood with side $4\,\frac{1}{\expFrac}|\traj{t-1}|$ and centered on the position of the previous localization. Note that $\frac{1}{\expFrac}|P_{t-1}|$, where $\expFrac$ is the exposure fraction, is the full trajectory length between $\frm{t-1}$ and $\frm{t}$. For each connected component in this region, the trajectory $\traj{t}$ and radius $\rad$ are computed in the same way as in the detection algorithm. The mean color $\col$ is obtained by averaging all pixel values on the trajectory. In this region, connected components with model parameters $(\col,\rad)$ are selected if the Euclidean distance in RGB $\|\col - \col_0\|_2$ and the normalized difference $\left|\rad - \rad_0\right|/\rad_0$ are below prescribed thresholds $0.3$ and $0.4$, respectively. Here, the previous FMO parameters are denoted by $(\col_0,\rad_0)$.

% In the beginning, re-detector operates on the same triple of frames as detector to produce $\Delta$, but cropped to the local neighborhood of the previous detection. Local neighborhood is defined as a square with the center in the centroid of the FMO and side length $4 (1 + \frac{1}{\expFrac}) \left| \traj{t} \right|$, where $(1 + \frac{1}{\expFrac}) \left| \traj{t} \right|$ is trajectory length with full exposure. In this region we search for objects that have $\col$ and $\rad$ consistent with the previous detection. If we define $\col_0$ and $\rad_0$ as parameters of newly found FMO, the consistency with the given model is defined as euclidean distance between RGB colors $|\col - \col_0|$ and normalized difference $\frac{\left|\rad_0 - \rad\right|}{\rad}$ to be less then a threshold (was set empirically). The knowledge of $\col, \rad, \traj{t-1}$ gives us sufficient constraints for robust seeking of $\traj{t}$. In addition, $\traj{t}$ is allowed to be constructed from several connected components\footnote{This was not acceptable in the detection due to false positives.}. Figure \ref{fig:examples} shows an example where the re-detector outperforms the detector. 

\input{figures/detection_fig}

\subsection{Tracker}
The final attempt to find the FMO after both the detector and re-detector have failed is the tracker, which uses image synthesis. The tracker is based on the simplified formation model \eqref{eq:model} by assuming an object $F$ with color $\col$ and radius $\rad$ moving along a linear trajectory $\traj{t}$. The indicator function $M$ is then a ball of radius $\rad$, and given the trajectory $\traj{t}$, the alpha value $[\degop_t M](x)$ from \eqref{eq:model} is
\begin{equation}
A(x|\traj{t}) = \frac{1}{|\traj{t}|} [\traj{t} \ast M](x) = \frac{1}{|\traj{t}|}\int_{|z| \leq r} \traj{t}(x-z) dz.
\label{eq:dist_tran}
\end{equation}
For linear trajectories this integral can be solved analytically. Let $D(x,\traj{t})$ denote the distance function from $x$ to the trajectory $\traj{t}$, then $A$ is 
\begin{equation}
A(x|\traj{t}) \approx \frac{2}{|\traj{t}|} \sqrt{\max\left( (\rad^2 - D^2(x,\traj{t}) ),0\right)}.
\end{equation}
This approximation is inaccurate only in the neighborhood of the starting and ending point of the trajectory, and for FMOs this area is small compared to the central section of the trajectory. Using the above relation, $\frm{t}$ in \eqref{eq:model} can be written in a simpler form as 
\begin{equation}
\hat{\frm{t}}(x|\traj{t}) =  (1 - A(x|\traj{t}))B(x) + \col A(x|\traj{t}).
\label{eq:mix}
\end{equation}

The tracker now looks for the trajectory $\traj{t}$ that best explains the frame $\frm{t}$ using the approximation $\hat{\frm{t}}$. This is equivalent to solving
\begin{equation}
\traj{t} = \arg \min_{\traj{t}} \|\hat{\frm{t}}(\cdot|\traj{t}) - \frm{t}\|_2.
\label{eq:loss}
\end{equation}
As in the other two algorithms, instead of the background $B$ we can use one of the previous frames $I_{t-1}$ or $I_{t-2}$, since a proper FMO should not occupy the same region in several consecutive frames, and thus the previous frame can locally serve as the background.

A linear trajectory $\traj{t}$ is given by its starting point $s_t$, orientation $\beta_t$ and length $|\traj{t}|$ (equivalently ending point $e_t$). We minimize \eqref{eq:loss} over these parameters by a coordinate descent search.

First, we find the best orientation. We extrapolate the starting point linearly from the previous detection and assume that the length remains the same, $s_t = e_{t-1} + \left(\frac{1}{\expFrac}-1\right)|P_{t-1}|u_{\beta_t}$ and $|P_t|=|P_{t-1}|$, where $u_{\beta}=\left(\cos(\beta),\,\sin(\beta)\right)$ is a unit vector with orientation $\beta$. Next we sample the space of $\beta_t$'s that differ from $\beta_{t-1}$ by up to 15$^\circ$ and choose the one that minimizes the cost \eqref{eq:loss}.

\begin{figure} [t]
\begin{tabular}{@{}c@{\hskip 1px}c@{\hskip 1px}c@{}}
 \includegraphics[width=0.33\linewidth]{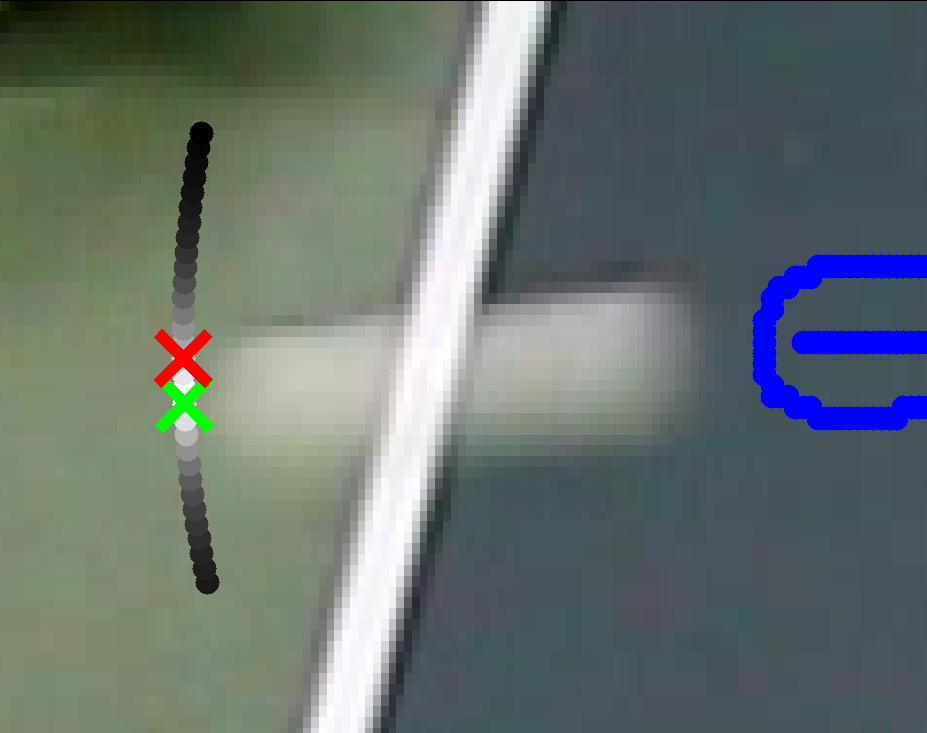} &
 \includegraphics[width=0.33\linewidth]{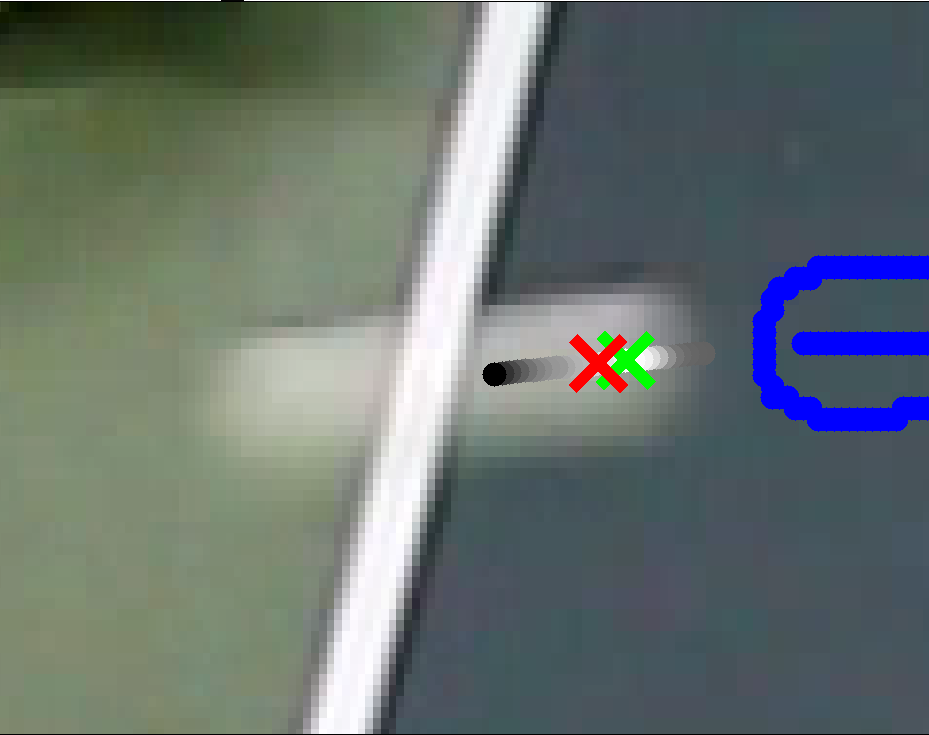} &
 \includegraphics[width=0.33\linewidth]{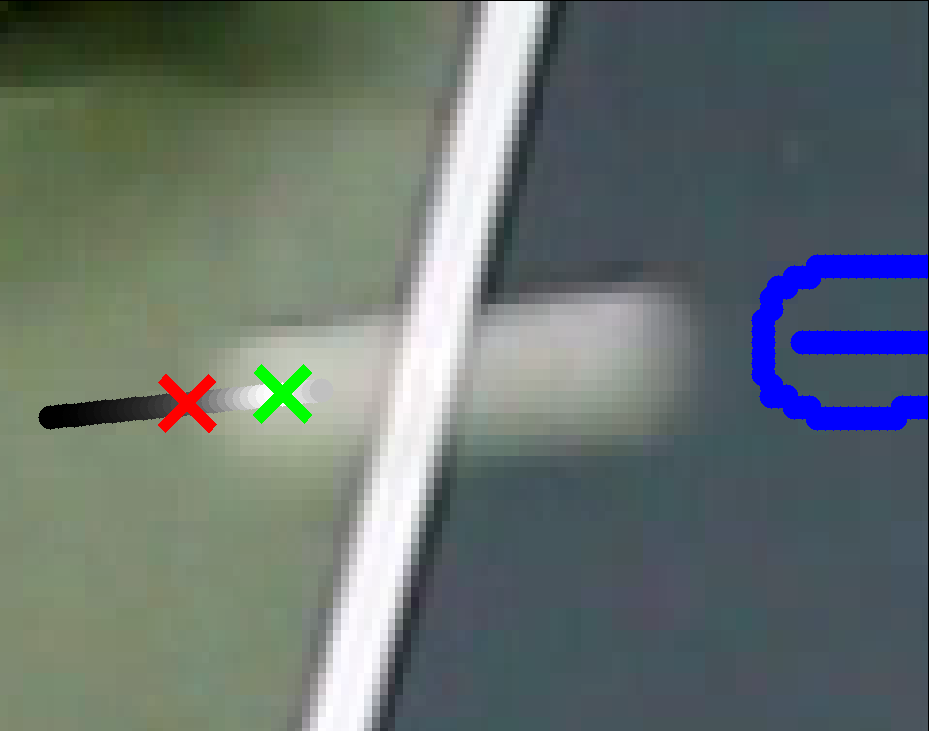} \\
 (1) & (2) & (3)
\end{tabular}
\caption{Tracking steps. (1) detection of orientation, (2) detection of starting point, (3) detection of ending point. Previous detection is in blue. Green cross denotes the minimizer, red crosses the initial guess. All sampled points (gray) are scaled by their cost \eqref{eq:loss} (the darker the higher cost).}
\label{fig:tracker}
\end{figure}

The minimization w.r.t. $s_t$ and $|\traj{t}|$ is done in a similar manner. For $s_t$, we sample points in the $\frac{1}{2}|\traj{t-1}|$ neighborhood of the extrapolated $s_t$ from the previous detection, and for $|P_t|$ we again use the range $|\traj{t-1}|\pm 50\%$. The three minimization stages are illustrated in Fig.~\ref{fig:tracker}.

\def\arraystretch{0.5}
\begin{figure*}
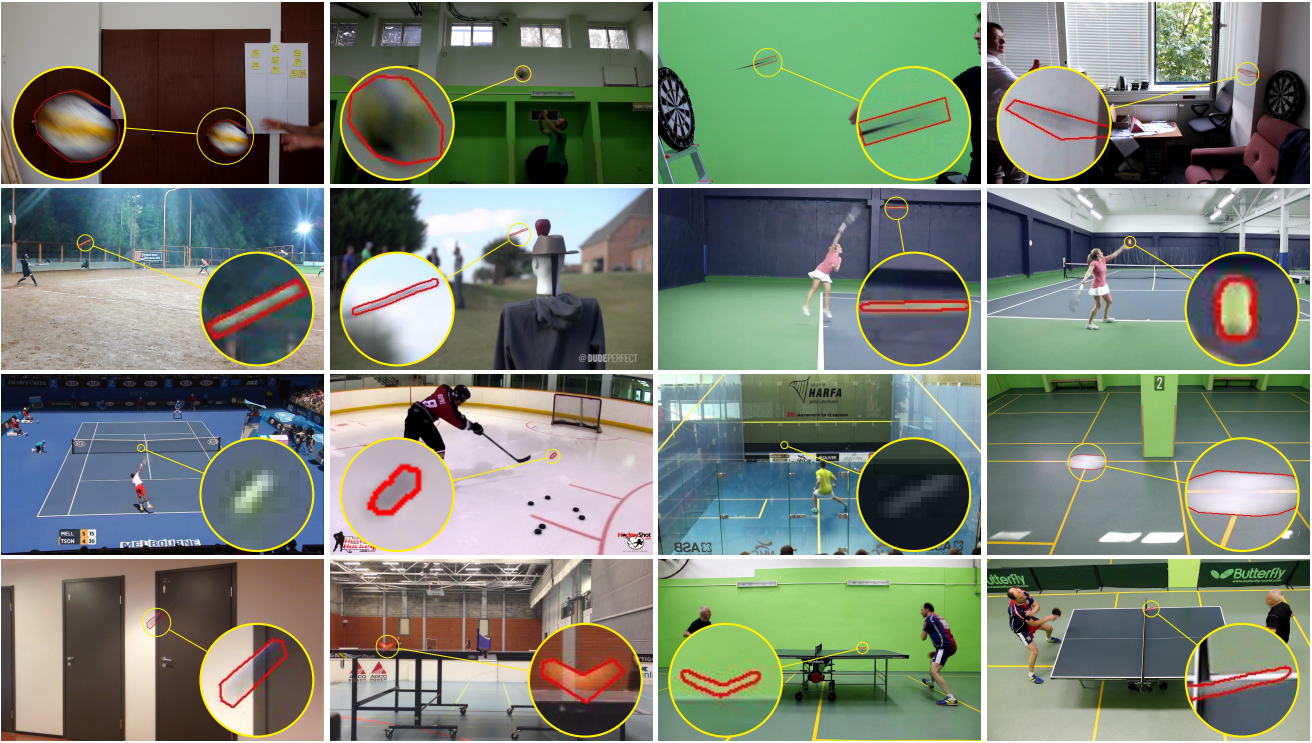

\centering
\begin{tabular}{@{}c@{\cscl}c@{\cscl}c@{\cscl}c@{\cscl}c@{}}
	\imgZoom{gt/volleyball1_gt}{\volleyballCoor}{2}{\defaultLeft} &
	\imgZoom{gt/volleyball_passing_gt}{\volleyballpassingCoor}{7}{\defaultLeft} &
	\imgZoom{gt/darts1_gt}{\dartsCoor}{4}{\defaultRight} &
	\imgZoom{gt/darts_window1_gt}{\dartswindowCoor}{5}{\defaultLeft} \\

	\imgZoom{gt/softball_gt}{\softballCoor}{7}{\defaultRight} &
	\imgZoom{gt/william_tell_gt}{\williamtellCoor}{5}{\defaultLeft} &
	\imgZoom{gt/tennis_serve_side_gt}{\tennisservesideCoor}{5}{\defaultRight} &
	\imgZoom{gt/tennis_serve_back_gt}{\tennisservebackCoor}{10}{\defaultRight} \\

	\imgZoom{seq/tennis1}{\tennisCoor}{16}{\defaultRight} &
	\imgZoom{gt/hockey_gt}{\hockeyCoor}{8}{\defaultLeft} &
	\imgZoom{seq/squash}{\squashCoor}{16}{\defaultRight} &
	\imgZoom{gt/frisbee_gt}{\frisbeeCoor}{3}{\defaultRight} \\

	\imgZoom{gt/blue_gt}{\blueCoor}{4}{\defaultRight} &
	\imgZoom{gt/ping_pong_paint_gt}{\pingpongpaintCoor}{5}{\defaultRight} &
	\imgZoom{gt/ping_pong_side_gt}{\pingpongsideCoor}{10}{\defaultLeft} &
	\imgZoom{gt/ping_pong_top_gt}{\pingpongtopCoor}{7}{\defaultRight} \\
\end{tabular}
\caption{The FMO dataset -- one example image per sequence. Red polygons delineate ground truth regions with fast moving objects. For clearer visualization two frames do not show annotations because their area consists only of several pixels. The sequences are sorted in natural reading order from left to right and top to bottom as in Tab. \ref{tab:performance}.}
\label{fig:dataset}
\end{figure*}
\def\arraystretch{1}

\section{Estimation of appearance} \label{sec:deconv}
%The object appearance can be estimated precisely when the FMO trajectory and size have been identified. 
%This for example allows us to slow down the object motion~--~basically increase the video frame rate~--~which may in some applications significantly contribute to the viewer experience. 
Let us consider a video frame $I_t$ acquired according to \eqref{eq:model} and the object trajectory $P_t$ and size $\rad$ as determined by the FMO detector. The objective is to estimate the appearance $F$, which is essentially a (modified) blind image deblurring task.
%, well studied in image processing and far from easy. 
One has to first estimate the blur-and-projection operator $\degop$, and then solve the non-blind deblurring task for $F$. 
%Estimating $\degop$ is undoubtedly the harder of the two steps, therefore, 
As mentioned in Sec. \ref{sec:problem}, to make the estimation of $\degop$ tractable, we focus on ball-like objects moving (approximately) parallel to the camera while undergoing arbitrary 3D rotation. As in the FMO tracking, if the object rotation is negligible or unperceivable, the $\degop$ operator is fully determined by the object trajectory and we can proceed directly to the non-blind estimation of $F$. Let us first focus on the estimation of $F$ and then on the problem of obtaining $\degop$.

%Once we have the blur operator $\degop$, we solve the nonblind step (estimating $F$) as follows. 
Let $F$ denote some representation of the object appearance~--~in the absence of rotation, this can be directly the image of the object projected in the video frame, and when 3D rotation is present we use the spherical parametrization to capture the whole surface. Following the model \eqref{eq:model} we solve the problem
\begin{equation}\label{eq:deblurring_nonblind}
	\min_F \|(1-[\degop M])B+[\degop F]-I\|_1 + \alpha \|\mathcal{D}F\|_1,
\end{equation}
where $\mathcal{D}$ is the derivative operator (gradient magnitude) and $\alpha$ is the weighting parameter proportional to the level of noise in $I$.  The $L_1$-norm while increasing robustness leads to nonlinear equations.  We therefore apply the method of iteratively re-weighted least squares to convert the optimization problem to a linear system and use conjugate gradients to solve it. For object sizes in the FMO dataset ($\rad<100$ pixels) this can be done in less than a second.

In the case of object rotation, the blur operator $\degop$ encodes the object pose (orientation in space) as well as location in each fractional moment during the camera exposure. Trajectory aside, this is fully determined by the object's angular velocity, which we assume constant throughout the exposure. Angular velocity (in 3D) is given by three parameters (two for axis orientation, one for velocity). The functional in \eqref{eq:deblurring_nonblind} is non-convex w.r.t the angular velocity parameters. However, we can solve it with an exhaustive search since the parametric space is not that large. We thus construct $\degop$ for each point in the discretized space of possible angular velocities, estimate $F$, and then measure the error given by the functional in \eqref{eq:deblurring_nonblind}. The parametrization which gives the lowest error is our solution.

\begin{figure*}
\centering
\begin{tabular}{@{}c|c@{}c@{}c@{}c@{}c@{}}
	\includegraphics[width=.16\textwidth]{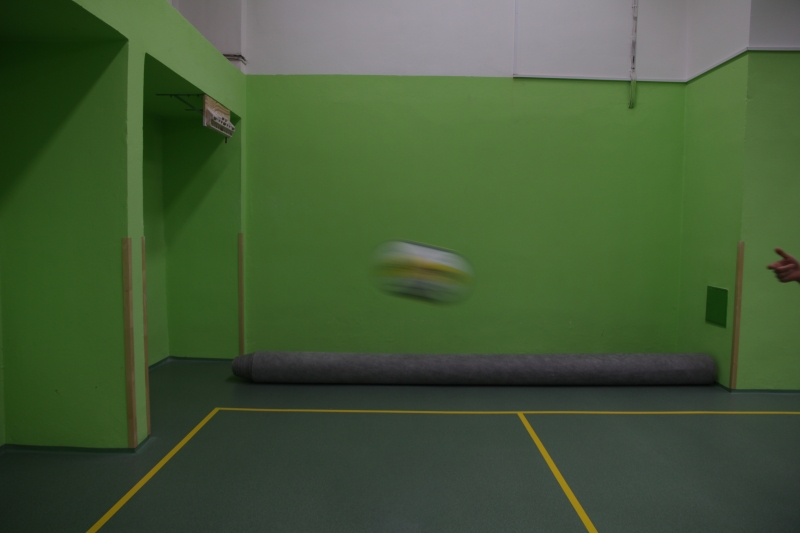} %
	&
	\drawImageAndCenters{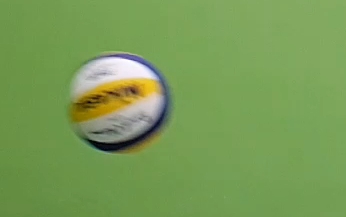}{0.158\textwidth}{346}{217}{109}{114}{113}{120} &
	\drawImage{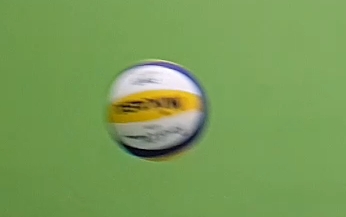}{0.158\textwidth} &
	\drawImage{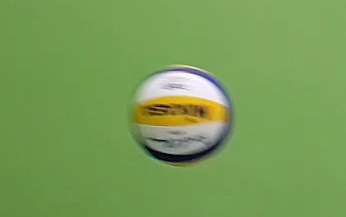}{0.158\textwidth} &
	\drawImage{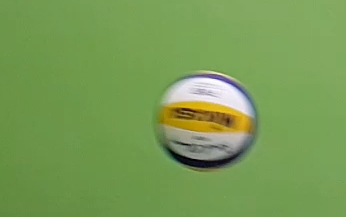}{0.158\textwidth}  &
	% \includegraphics[width=.16\textwidth]{img/ballRotation/volleyball_05/02_gt} \hfill %
	% \drawImageAndCenters{img/ballRotation/volleyball_05/01_gt}{0.158\textwidth}{346}{217}{300}{100}{300}{100} \\
	\drawImage{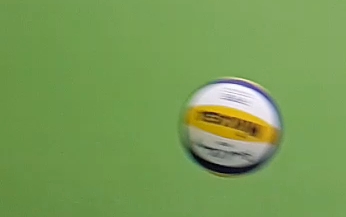}{0.158\textwidth} \\ %
	\includegraphics[width=.16\textwidth]{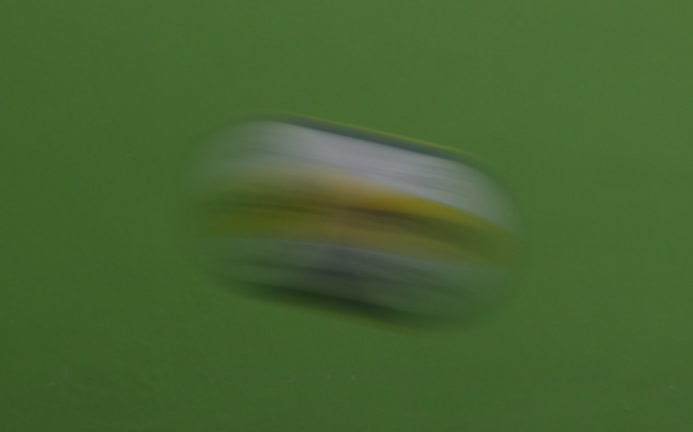} %
	&
	\drawImage{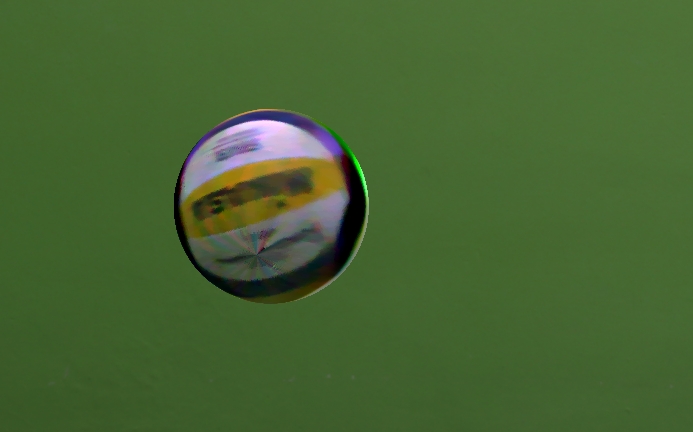}{0.158\textwidth}  &
	\drawImage{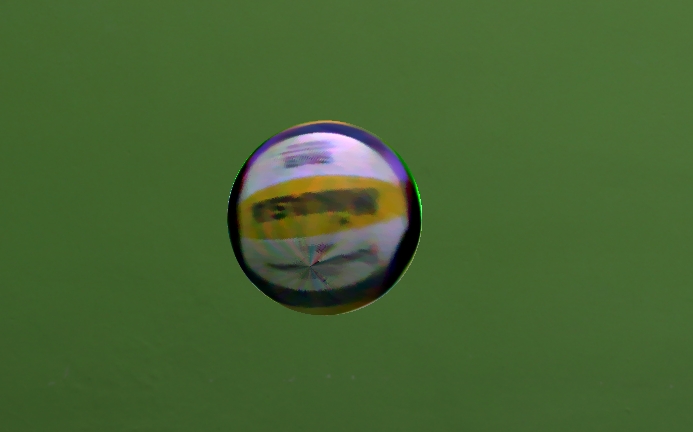}{0.158\textwidth}  &
	\drawImage{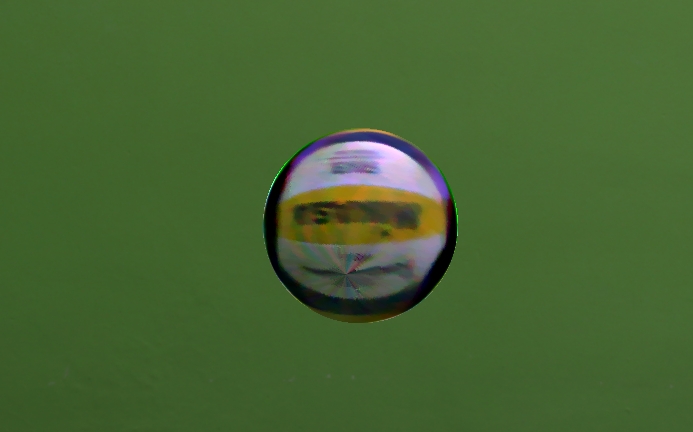}{0.158\textwidth}  &
	\drawImage{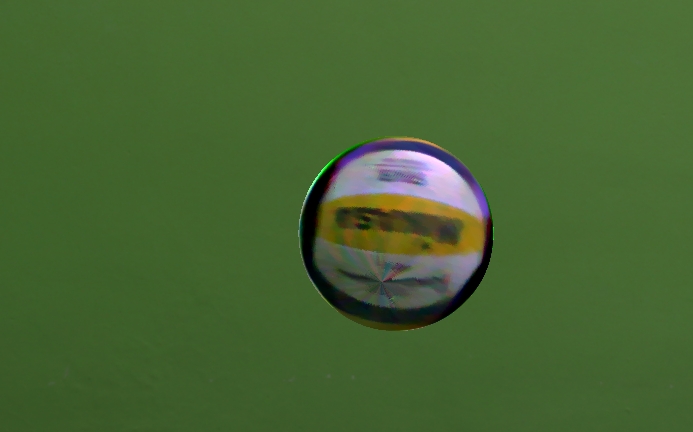}{0.158\textwidth}  &
	\drawImage{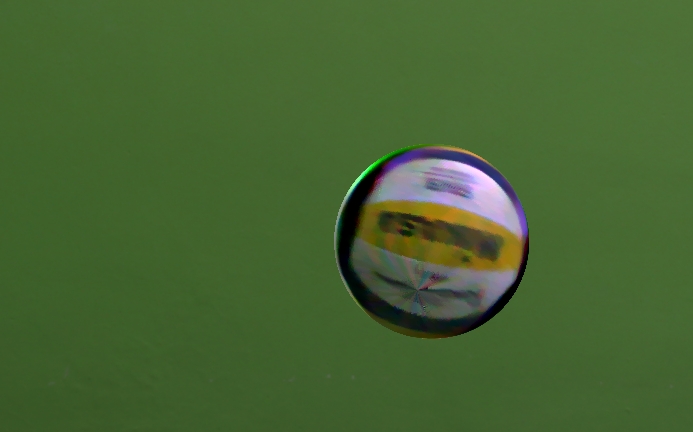}{0.158\textwidth} \\ %
	\multicolumn{1}{c}{a)} & b)
\end{tabular}
\caption{Reconstruction of an FMO blurred by motion and rotation. a) Input video frame. b) Top row: actual frames from a high-speed camera (250fps). Bottom row: frames at corresponding times reconstructed from a single frame of a regular camera (25fps), i.e. 10x temporal super-resolution. The top left image shows the rotation axis position estimated from the blurred frame (blue cross) and from the highspeed video (red circle).}
\label{fig:volleyball_SR}
\end{figure*}

In Fig.~\ref{fig:volleyball_SR} we illustrate the result of FMO deblurring in the form of temporal super-resolution. The left side (a) shows a frame captured by a conventional video camera (25fps), which contains a volleyball that is severely motion blurred. On the right side (b), the top row shows several frames captured by a high-speed video camera (250fps) spanning approximately the same time frame~--~the volleyball flies from left to right while rotating clockwise. In the bottom row of (b) we show the result of FMO deblurring, computed solely from the single frame in (a), at times corresponding to the high-speed frames above. The restoration is on par with the high-speed ground-truth; it significantly enhances the video information content merely by post-processing. For comparison, we also display the calculated rotation axis and the one estimated from the high-speed video. Both are close to each other; compare the blue cross and red circle in (b). Note that for a human observer it is impossible to determine the ball rotation from blurred images while the proposed algorithm with the temporal super-resolution output provides this insight. Another appearance estimation example is in Fig.~\ref{fig:superres}, where we use the simplified model of pure translation motion for the table-tennis ball (top) and frisbee (bottom).

%1) Formulate the problem of parameter estimation as the minimization of functional

% 2) Point out that the ``no-blur'' solution that is a common problem in MAP blind deconvolution does not occur here.  The reason is the presence of occlusion in \eqref{eq:model}.

% 3) Describe briefly the algorithm. Initialization of the trajectory $\tau_i$ and object mask $m$ from tracking. Alternating minimization with respect to $f$, $m$ and $\tau_t$. Minimization with respect to the rotation parameters $a_t$ and $\phi_t$ must be done as full search....

\newcommand{\moreResVar}{0.49}
\def\arraystretch{0.5}
\begin{figure}[t]
\centering
\begin{tabular}{@{}c@{\hskip 1.5px}c@{}}
	\includegraphics[width=\moreResVar\linewidth]{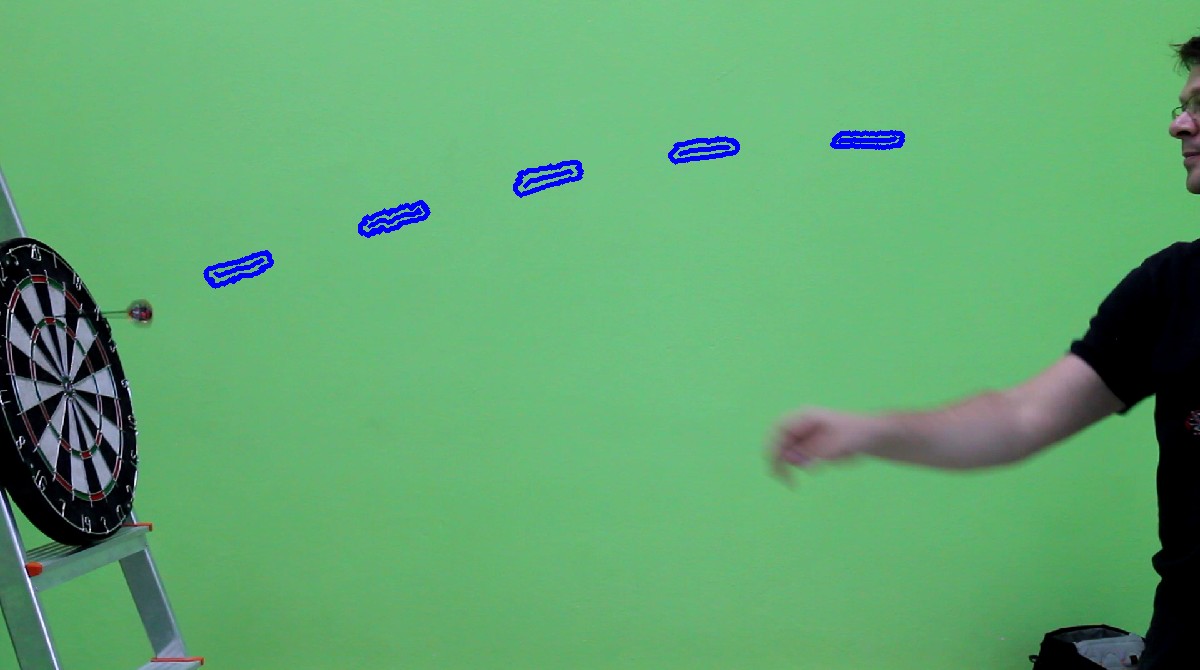} & 
	\includegraphics[width=\moreResVar\linewidth]{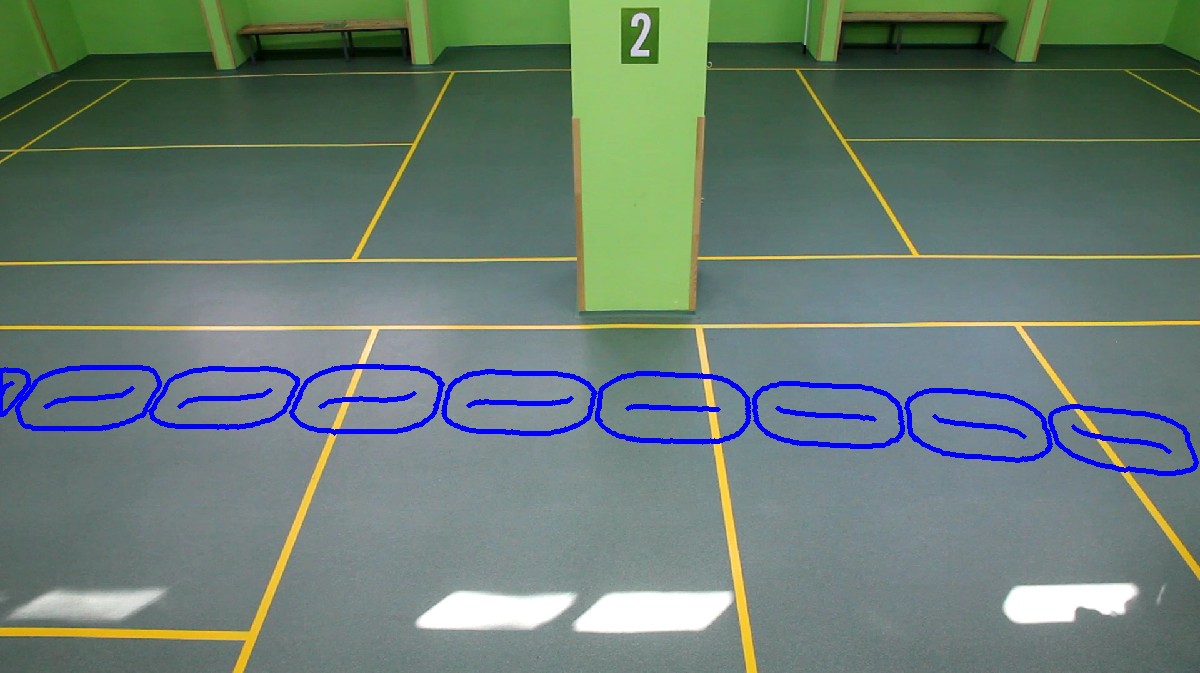} \\
	\includegraphics[width=\moreResVar\linewidth]{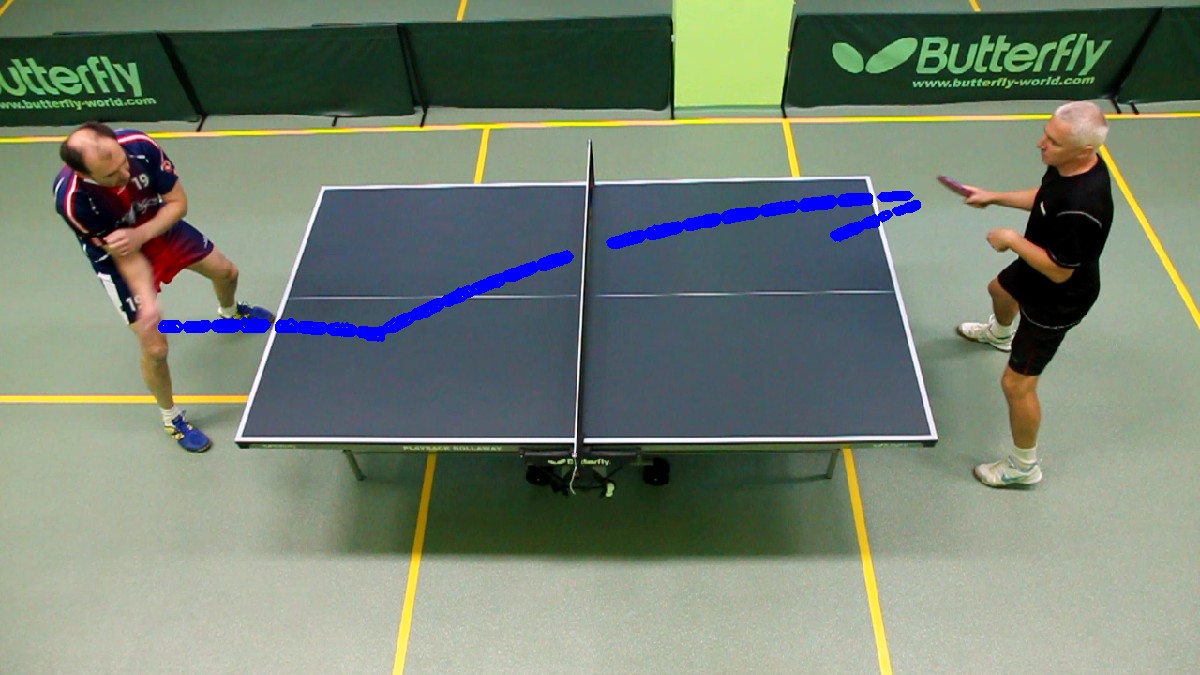} &
	\includegraphics[width=\moreResVar\linewidth]{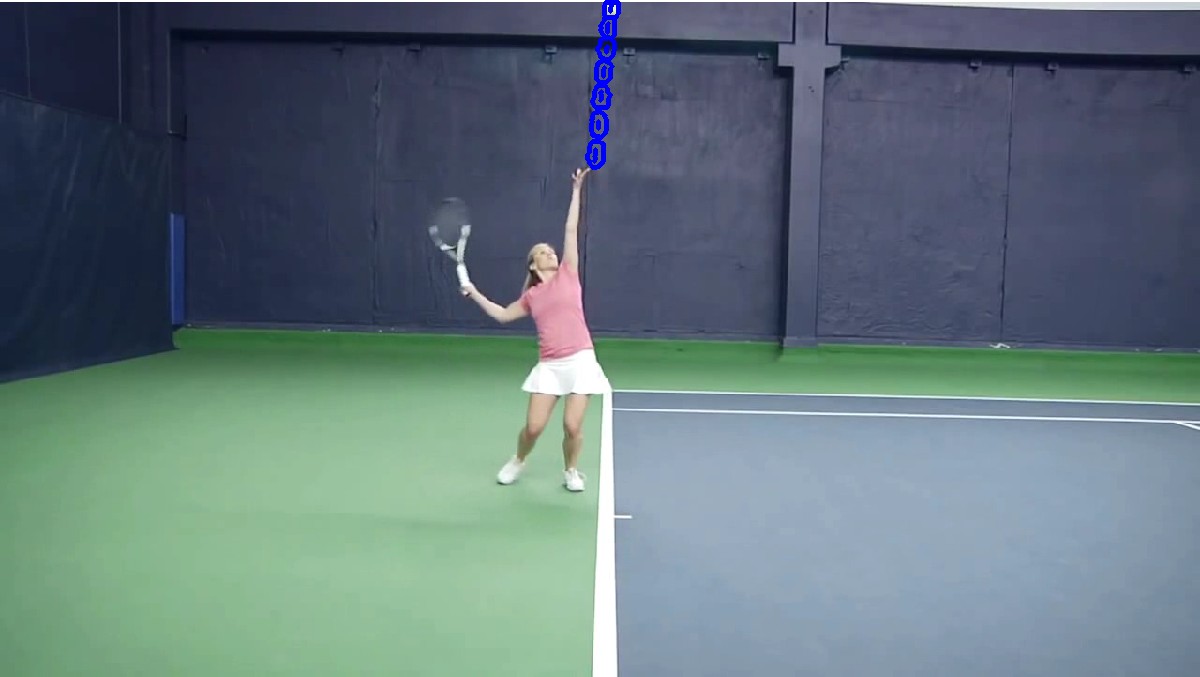} \\
\end{tabular}
\caption{FMO detection and tracking. Each blue region refers to the object trajectory and contour in previous frames.}
\label{fig:more_results}
\end{figure}
\def\arraystretch{1}

\section{Dataset}  \label{sec:dataset}
The FMO dataset contains videos of various activities involving fast moving objects, such as ping pong, tennis, frisbee, volleyball, badminton, squash, darts, arrows, softball, as well as others. Acquisition of the videos differ: some are taken from a tripod with mostly static backgrounds, some have severe camera motions and dynamic backgrounds, some FMOs are nearly homogeneous, while some have colored texture. All the sequences are annotated with ground-truth locations of the object (even in cases when the object of interest does not strictly satisfy the notion of FMO).

None of the public tracking datasets contain objects moving fast enough to be considered FMOs~--~with significant blur and large frame-to-frame displacement. We analyzed three of the most widely used tracking datasets, ALOV \cite{alov}, VOT \cite{vot2015}, and OTB \cite{otb} and compared them with the proposed method in terms of the motion of the object of interest. For example, in the conventional datasets, the object frame-to-frame displacement is below 10 pixels in 91\% of cases, while in the FMO dataset the displacement is uniformly spread between 0 and 150 pixels. Similarly, the intersection over union (IoU) of bounding boxes between adjacent frames is above $0.5$ in 94\% of times for the conventional datasets, whereas the proposed dataset has zero intersection nearly every time. Fig.~\ref{fig:hist} summarizes these findings. 

An overview of the FMO dataset is in Fig.~\ref{fig:dataset}, showing some of the included activities and the ground-truth annotations. The dataset and annotations will be made publicly available.

\begin{table}[t]
\begin{tabular}{r|l||r|r|r|r}
\hline
n & Sequence name & \# & Pr. & Rc. & F-sc. \\
\hline
1&volleyball         & 50 & 100.0& 45.5 & 62.5 \\
2&volleyball passing & 66 & 21.8 & 10.4 & 14.1 \\
3&darts 			   & 75 & 100.0& 26.5 & 41.7 \\
4&darts window       & 50 & 25.0 & 50.0 & 33.3 \\
5&softball           & 96 & 66.7 & 15.4 & 25.0 \\
6&archery       & 119& 0.0  & 0.0  & 0.0  \\
7&tennis serve side  & 68 & 100.0& 58.8 & 74.1 \\
8&tennis serve back  & 156& 28.6 & 5.9  & 9.8  \\
9&tennis court       & 128& 0.0  & 0.0  & 0.0  \\
10&hockey             & 350& 100.0& 16.1 & 27.7 \\
11&squash             & 250& 0.0  & 0.0  & 0.0  \\
12&frisbee            & 100& 100.0& 100.0& 100.0\\
13&blue ball          & 53 & 100.0& 52.4 & 68.8 \\
14&ping pong tampere  & 120& 100.0& 88.7 & 94.0 \\
15&ping pong side     & 445& 12.1 & 7.3  & 9.1  \\
16&ping pong top      & 350& 92.6 & 87.8 & 90.1 \\
\hline
&Average & -- & 59.2 & 35.5 & 40.6 \\
\hline
\end{tabular}
\caption{Performance of the proposed method on the FMO dataset. We report precision, recall and F-score. The number of frames is indicated by \#.}
\label{tab:performance}
\end{table}
\begin{table}[t]
\begin{tabular}{l||r|r|r|r|r|r}
\hline
\diagbox{Sq. name}{\rotatebox[origin=c]{90}{Method}} & \rotatebox[origin=c]{90}{ASMS\cite{asms}} & \rotatebox[origin=c]{90}{DSST\cite{dsst}} & \rotatebox[origin=c]{90}{MEEM\cite{meem}} & \rotatebox[origin=c]{90}{SRDCF\cite{srdcf}} & \rotatebox[origin=c]{90}{STRUCK\cite{struck}}& \rotatebox[origin=c]{90}{Proposed} \\
\hline
volleyball         & \textbf{80} & 0 & 50 & 0 & 10 & 46\\
volleyball passing & 12 & 6 & \textbf{95} & 88 & 8 & 10 \\
darts 			   & 3 & 0 & 6 & 0 & 0 & \textbf{27} \\
darts window       & 0 & 0 & 0 & 0 & 0 & \textbf{50} \\
softball           & 0 & 0 & 0 & 0 & 0 & \textbf{15} \\
archery       & \textbf{5} & \textbf{5} & \textbf{5} & \textbf{5} & 0 & 0 \\
tennis serve side  & 7 & 0 & 0 & 0 & 6 & \textbf{59} \\
tennis serve back  & 5 & 0 & 0 & 0 & 3 & \textbf{6} \\
tennis court       & 0 & 0 & \textbf{3} & \textbf{3} & 0 & 0 \\
hockey             & 0 & 0 & 0 & 0 & 0 & \textbf{16} \\
squash             & 0 & 0 & 0 & 0 & 0 & 0 \\
frisbee            & 65 & 0 & 6 & 6 & 0 & \textbf{100} \\
blue ball          & 30 & 0 & 0 & 0 & 25 & \textbf{52} \\ 
ping pong tampere  & 0 & 0 & 0 & 0 & 0 & \textbf{89} \\
ping pong side     & 1 & 0 & 0 & 0 & 0 & \textbf{7} \\
ping pong top      & 0 & 0 & 0 & 0 & 1 & \textbf{88} \\
\hline
Average & 17 & 1 & 1 & 1 & 3 & \textbf{36} \\ 
\hline
\end{tabular}
\caption{Performance of baseline methods on the FMO dataset. Percentage of frames with FMOs present where tracking was successful (IoU $>$ $0.5$).}
\label{tab:baselines}
\end{table}

\def\arraystretch{0.5}
\def\arraystretch{1}
\section{Evaluation}  \label{sec:eval}
The proposed localization pipeline was evaluated on the FMO dataset.
The performance criteria are precision $\textrm{TP}/(\textrm{TP}+\textrm{FP})$, recall $\textrm{TP}/ (\textrm{TP}+\textrm{FN})$ and F-score $2 \textrm{TP}/(2 \textrm{TP} + \textrm{FN } + \textrm{FP})$, where TP, FP, FN is the number of true positives, false positive and false negatives, respectively. A true positive detection has an intersection over union (IoU) with the ground truth polygon  greater than 0.5 and  an IoU larger than other detections. The second condition ensures that multiple detections of the same object generates only one TP.  False negatives are FMOs in the ground truth with no associated FP detection. 

Quantitative results for individual video sequences are listed in Tab.~\ref{tab:performance}. 
All results were achieved for the same set of parameters in the localization pipeline as discussed in Sec.~\ref{sec:method}. 
Performance varies widely, ranging from a F-score of $0\%$ (complete failure) for the archery, tennis court, and squash sequences, to $100\%$ (complete success) for the frisbee sequences. 
The sequences with the best results contain objects with prominent FMO characteristics, i.e. a large motion against a contrasting background.  
False negatives occur in three types of situations: (i) the object motion is too small (archery, volleyball), (ii) the object itself is too small (tennis court, squash), and (iii) the background is too similar to the object color (e.g., table tennis net, white edge of the table). Problem (i) can be addressed by combining the FMO detector with a state-of-the-art ``slow'' short-term tracker. False positives usually occur when local movements of larger objects, such as players' body parts, can be partially explained by the FMO model, or due to imprecise camera stabilization. Note that none of the test sequences contain multiple FMOs in a single frame, but the algorithm is not constrained to detect a fixed number of objects. The detection results are included in the supplementary material. Some examples are shown in Fig.~\ref{fig:more_results}.

Next, we compare the results of the FMO localization pipeline to those of several standard state-of-the-art trackers, namely ASMS \cite{asms}, DSST \cite{dsst}, SRDCF \cite{srdcf}, MEEM \cite{meem}, and STRUCK \cite{struck}. For a fair comparison, only frames containing exactly one FMO were included. Since these trackers always output exactly one detection per frame and the proposed method can return any number of detections, including none, the proposed method would have an advantage on the full set of frames.
The results are presented in Tab.~\ref{tab:baselines} in terms of the percentage of frames with a successful detection. Some of the standard trackers performed reasonably well on the volleyball sequences, where the motions are relatively slow, but overall results are very poor. The proposed method performs significantly better. This is explainable because the compared methods were not designed for scenarios involving FMOs, but it highlights the need for a specialized FMO tracker.

Besides FMO localization, the proposed model and estimator enable several applications which may be useful in processing videos containing FMOs. In Sec. \ref{sec:deconv}
on appearance estimation, we suggested the task of temporal super-resolution, which increases the video frame-rate by filling out the gap between existing frames and artificially decreases the exposure period of existing frames.
The naive approach is the interpolation of adjacent frames, which is inadequate for videos containing FMOs. A more precise approach requires moving objects to be localized, deblurred, and their motions modeled, which the proposed method accomplishes (see Sec.~\ref{sec:deconv}), so that new frames can be synthesized at the desired frame-rate. Figs.~\ref{fig:volleyball_SR} and \ref{fig:superres} show example results of the temporal super-resolution.

Another popular use case is highlighting FMO in sport videos. Due to the extreme blur, FMOs are often hard to localize, even for humans, despite having the context provided by perfect semantic scene understanding. 
Simple highlighting, like recoloring or scaling, enhances the viewer's experience. Fig.~\ref{fig:superres} top-right demonstrates temporal super-resolution with highlighting.

\let\thefootnote\relax\footnote{\textbf{Acknowledgments.} The authors were supported by the Technology Agency of the Czech Republic project TE01020415 V3C, the MSMT LL1303 ERC-CZ project and the Grant Agency of the Czech Republic under project GA13-29225S.}

\newcommand{\supHeight}{0.49}
\newcommand{\supHskip}{\hskip 1.5px}
\def\arraystretch{0.5}
\begin{figure}
\centering
\begin{tabular}{@{}c@{\supHskip}c@{}}
	\includegraphics[width=\supHeight\linewidth]{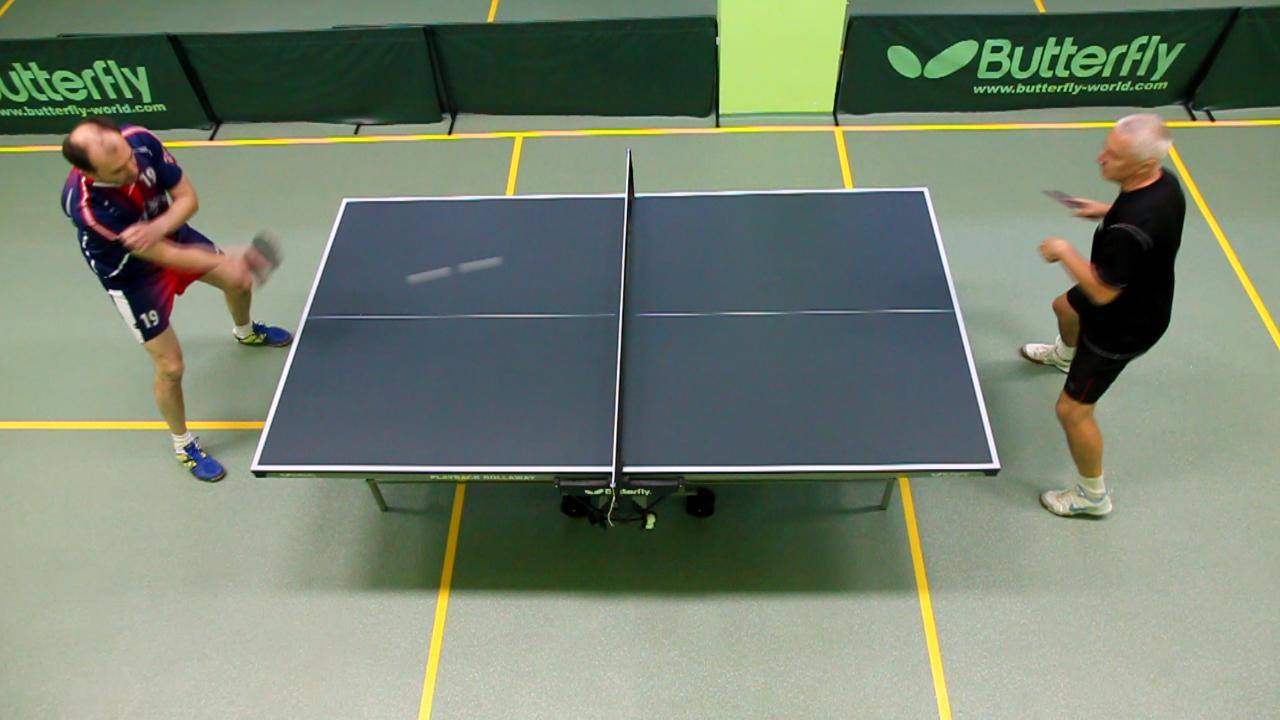} &
	\includegraphics[width=\supHeight\linewidth]{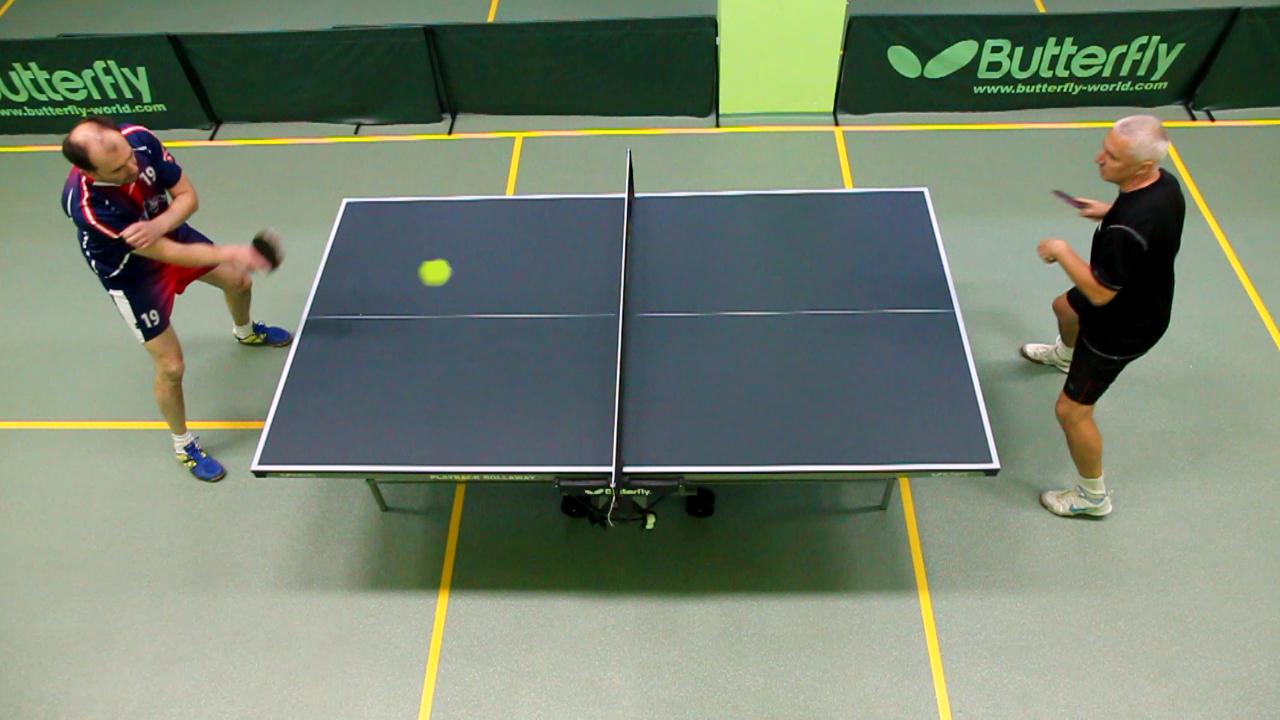} \\
	\includegraphics[width=\supHeight\linewidth]{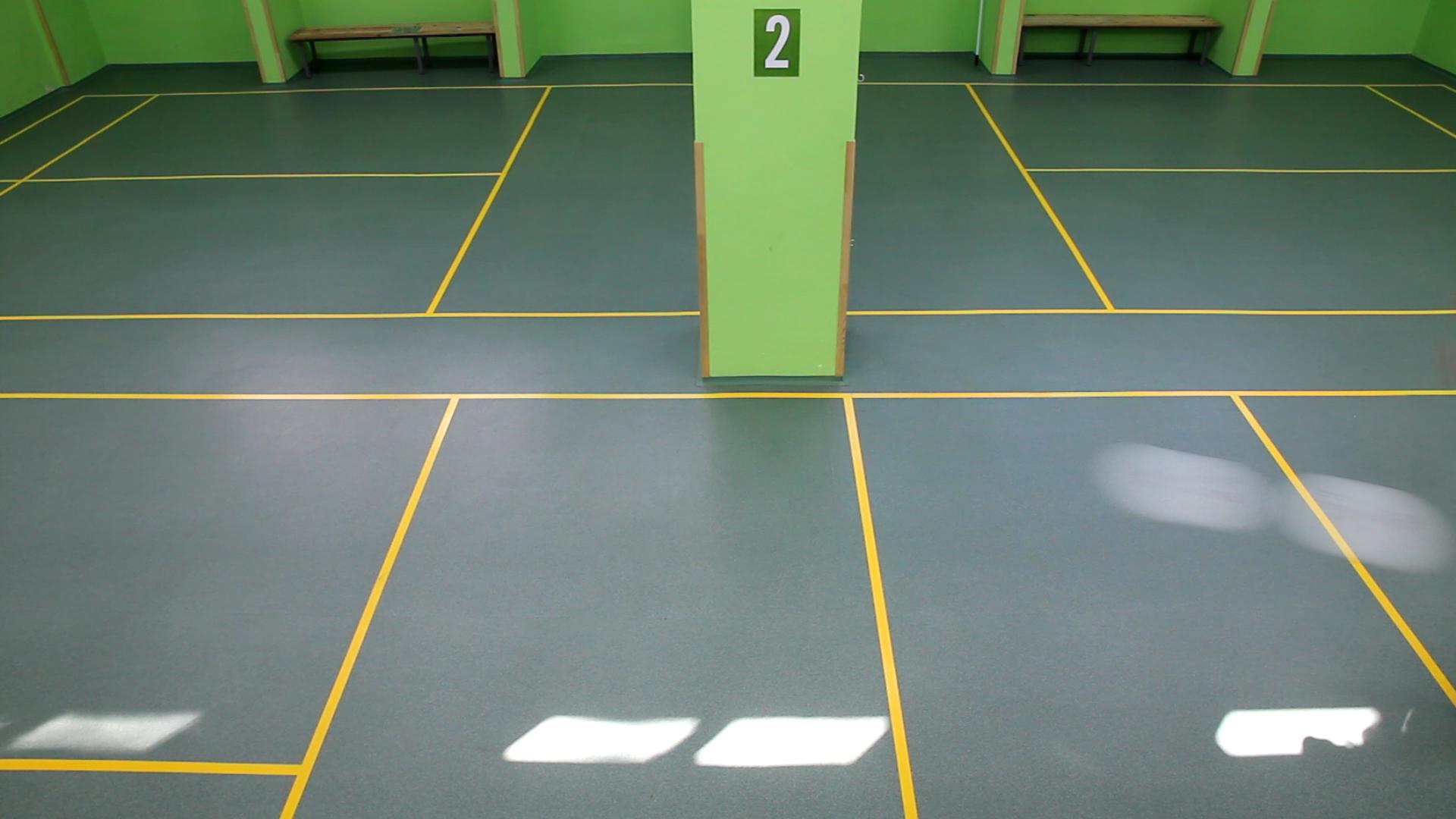} &
	\includegraphics[width=\supHeight\linewidth]{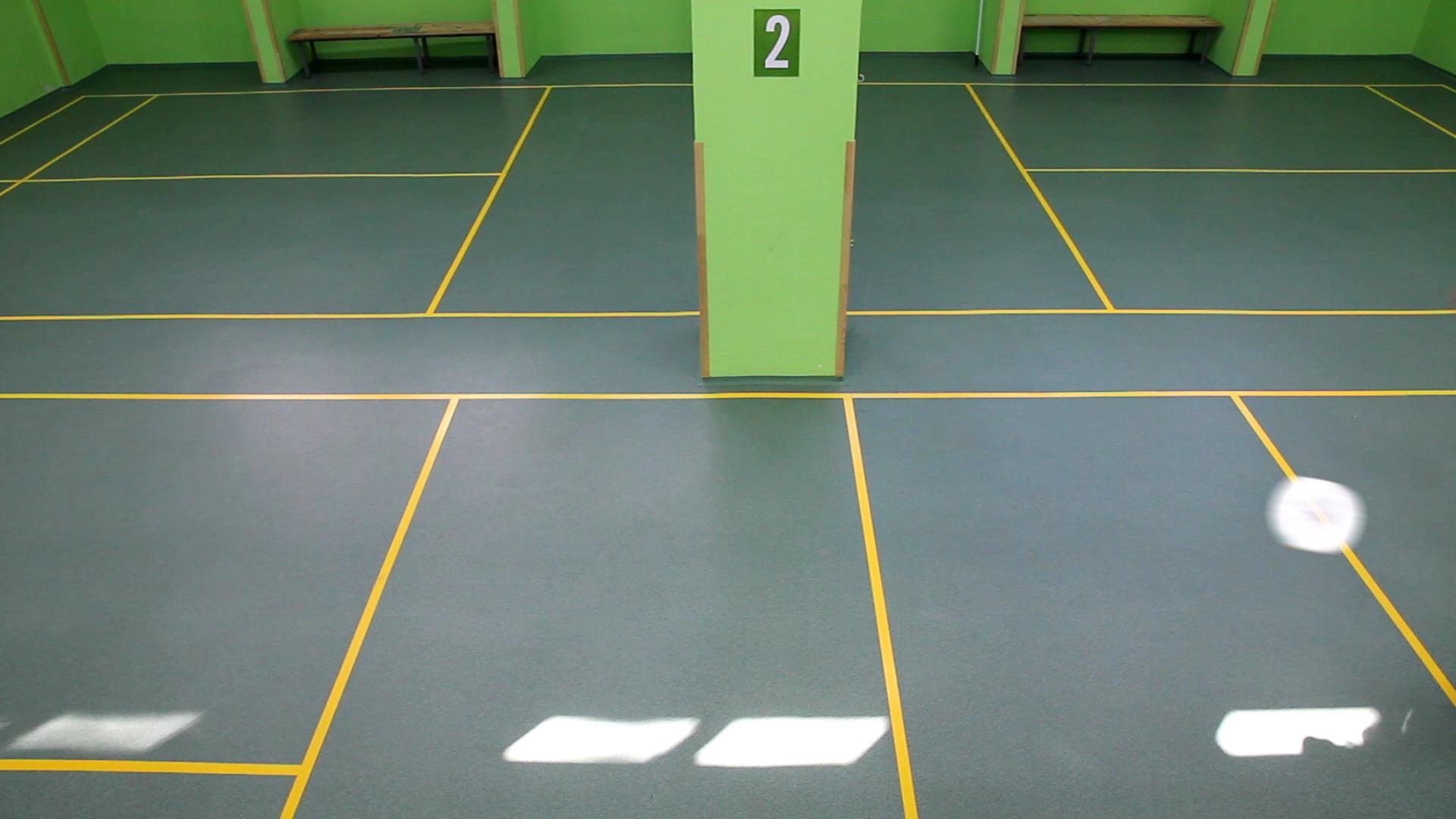} \\
\end{tabular}
\caption{Temporal super-resolution using plain interpolation (left) and the appearance estimation model (right). The top right image shows the possibility of FMO highlighting.}
\label{fig:superres}
\end{figure}
\def\arraystretch{1}

% \newcommand{\supHeight}{0.123}
% \newcommand{\supHskip}{\hskip 1.5px}
% \def\arraystretch{0.5}
% \begin{figure*}
% \centering
% \begin{tabular}{@{}c@{\supHskip}c@{\supHskip}c@{\supHskip}c@{\supHskip}c@{\supHskip}c@{\supHskip}c@{\supHskip}c@{}}
% 	\includegraphics[width=\supHeight\textwidth]{apps/average1} &
% 	\includegraphics[width=\supHeight\textwidth]{apps/average2} &
% 	\includegraphics[width=\supHeight\textwidth]{apps/average3} &
% 	\includegraphics[width=\supHeight\textwidth]{apps/average4} &
% 	\includegraphics[width=\supHeight\textwidth]{apps/average5} &
% 	\includegraphics[width=\supHeight\textwidth]{apps/average6} &
% 	\includegraphics[width=\supHeight\textwidth]{apps/average7} &
% 	\includegraphics[width=\supHeight\textwidth]{apps/average8} \\

% 	\includegraphics[width=\supHeight\textwidth]{apps/average_fmo1} & 
% 	\includegraphics[width=\supHeight\textwidth]{apps/average_fmo2} &
% 	\includegraphics[width=\supHeight\textwidth]{apps/average_fmo3} &
% 	\includegraphics[width=\supHeight\textwidth]{apps/average_fmo4} &
% 	\includegraphics[width=\supHeight\textwidth]{apps/average_fmo5} &
% 	\includegraphics[width=\supHeight\textwidth]{apps/average_fmo6} &
% 	\includegraphics[width=\supHeight\textwidth]{apps/average_fmo7} &
% 	\includegraphics[width=\supHeight\textwidth]{apps/average_fmo8} \\
% \end{tabular}
% \caption{Temporal superresolution by $8x$. Top row: using plain interpolation. Bottom row: interpolation with FMOs.}
% \label{fig:superres}
% \end{figure*}
% \def\arraystretch{1}
\section{Conclusions} \label{sec:conclusion}
Fast moving objects are a common phenomenon in real-life videos, especially sports. We proposed a generic, i.e. not requiring prior knowledge of appearance, algorithm for their fast localization and tracking and a blind deblurring algorithm for estimation of their appearance. 
We created a new dataset consisting of 16 sports videos with ground-truth annotations. Tracking FMOs is considerably different from standard object tracking targeted by state-of-the-art algorithms and thus requires a specialized approach. The proposed method is the first attempt in this direction and outperforms baseline methods by a wide margin. 
%pushes the baseline performance by a wide margin on this dataset and achieved high performance while showing that ordinary trackers are not capable of reliable FMO detection, a new approach is needed. 
The estimated FMO appearance could support applications useful in sports analytics, such as realistic increase of video frame-rate (temporal super-resolution), artificial object highlighting,  visualization of rotational axis and measurement of speed and angular velocity. 
%We also presented several useful applications~--~a temporal super-resolution of videos and FMO highlighting or visualization.

{\small
\bibliographystyle{ieee}
\bibliography{egbib}
}

\end{document}